\documentclass{article}
\usepackage[doublespacing]{setspace}
\usepackage[applemac]{inputenc}

\begin{document}

\title{Complex Networks}
\author{Carlos Gershenson$^{1,2}$ and Mikhail Prokopenko${^3}$  \\
$^{1}$Instituto de Investigaciones en Matem\'aticas Aplicadas y en Sistemas\\
Universidad Nacional Aut\'onoma de M\'exico\\
A.P. 20-726, 01000 M\'exico D.F. M\'exico\\
$^{2}$Centro de Ciencias de la Complejidad, UNAM, M\'exico\\
$^{3}$CSIRO Information and Communications Technology Centre,\\
Locked Bag 17, North Ryde, NSW 1670, Australia \\
}

\maketitle



Graph theory was initiated by Euler in the eighteenth century. In mathematics, graph theory consolidated itself in the following decades and centuries. However, it was only until a little more than a decade ago that an explosion of research and applications ocurred, in what is now referred to as network science \cite{Watts:1998,Barabasi2002,NewmanEtAl2006,Newman:2010}. In particular, the networks have become a central tool in the study of complex systems \cite{Bar-Yam1997,Mitchell:2006,Prokopenko:2008,Mitchell:2009}. The language of ``nodes and edges" provided  by networks has proven to be very illustrative to model elements of a system (nodes) and their interactions (edges). Having a language that describes interactions is essential for a non-reductionist science \cite{GershensonHeylighen2005}.

The relevance of the study complex networks \cite{Newman:2003,Caldarelli:2007,Barrat:2008,Cohen:2010} resides in the fact that so many real networks do not have a trivial topology. It follows that their properties are also not trivial, opening many research avenues. Examples of these properties include the small world effect, scale-free topologies, modularity, robustness, evolvability, degeneracy, and redundancy. 
 
Within ALife, almost all topics benefit from the study of complex networks, since the connectivity of systems is strongly related to their function. For example, cortical networks, genetic regulatory networks, metabolic pathways, artificial chemistries, and ecological webs describe phenomena in terms of nodes and links with a non-trivial topology. 

For this reason, we decided to organize a special session on complex networks at the ALife XII conference in Odense, Denmark, which was held on August $20^{th}$ and $21^{st}$, 2010. The intention of the session was to foster cross-fertilization between the ALife and complex networks communities. Following the success of the session, a call for papers for this special issue was launched.

We received fifteen submissions, out of which eight papers were selected with the valuable aid of multiple thorough reviews.

A generic unifying framework for diverse complex real-world networks has not yet been developed, and in part this is due to a limited number of available examples of these networks. As pointed out by Liu et al. \cite{liu} this can be addressed by development of re-wiring algorithms capable of generating networks with specific characteristics. Such characteristics may, for example, combine scale-free properties and community structures encountered in the real-world. The re-wiring algorithm presented in this work is inspired by observations of social interactions, capturing an appropriately tuned local-global coupling. The approach is verified by computational experiments, resulting in generation of networks that resemble their real-world counterparts in terms of important topological details.

Brede presents another model of network generation \cite{brede1}, where the rates of random addition of nodes and optimal rewiring are explored to generate complex networks with power law tails in degree distributions, hierarchies, non-trivial clustering and degree mixing patterns. 

Another step towards a generic framework for networks science is made by Lizier et al. \cite{lizier}, who investigate computational capabilities of small-world networks in terms of information-theoretic measures. The analysis includes topological and dynamical phase transitions, and associates specific modes of computation (such as storage and transfer) with well-known phases of the dynamics and randomness in topology. The main result is the observation that the information storage and information transfer are somewhat balanced near the small-world regime, providing quantitative evidence that small-world networks are capable of supporting comparably large information storage and transfer capacity. This observation may open a way for explaining, within a general information-theoretic framework, the prevalence of small-world occurrence in naturally occurring networks. 

On the same path lies the study of the order-chaos phase transition in random Boolean networks (RBNs), which have been used as models of gene regulatory networks \cite{Kauffman1969}, carried out by Wang et al. \cite{wang}. This work characterizes the RBN dynamics via a phase diagram obtained with respect to Fisher information. This novelty offers a natural interpretation of the phase diagram---through a generic measure capturing information-theoretically how much system dynamics can reveal about its control parameters. The observation that this measure is maximized near the order-chaos phase transitions concurs well with the characterization of the ``edge of chaos'' as the region where the system is most sensitive to its parameters. Importantly, the study exemplifies how Fisher information may be used as a powerful generic tool in network science.

Another paper that uses RBNs is that of Poblanno-Balp and Gershenson \cite{poblanno}, where the effect of a modular topology is studied on the properties and dynamics of RBNs. Most RBN studies have been made on homogeneous or normal topologies. However, it is known that real genetic regulatory networks have a modular structure. A general model of modular RBNs is presented. Statistical and analytical studies show that modularity can considerably change the properties of RBNs. In particular, modularity produces critical dynamics in networks where their average connectivity would suggest chaotic dynamics. Also, more attractors are observed on average in modular RBNs.

Droop and Hickinbotham \cite{droop} use an artificial chemistry to study the properties of networks constructed from mutation patterns observed in nature. The resulting small-world networks offer a balance between random and regular topologies, resonating with the results of Lizier et al. \cite{lizier}. This balance is advantageous for the exlporation of evolutionary space, i.e. evolvability. 

A second contribution by Brede \cite{brede2} studies the popular prisoner's dilemma \cite{Nowak2006} on regular and heterogenous networks with heterogeneous payoff landscapes. Results illustrate the non-trivial relations between network topology and the facilitation of the evolution of cooperation.

The final paper of the issue \cite{adami} presents a study of complex networks that utilizes colored motifs as the building blocks of the networks. Again, the approach turns to information theory as a generic method of choice: the motifs are used to define the information content of the network. Importantly, the colored motif information is related to aspects of selection, investigated via the interaction between instructions in genomes of digital life organisms, as well as the \emph{C. elegans} brain.  The central observation (that the colored motif information content changes during evolution, depending on how the genomes are organized) offers an interesting tool to dissect genomic rearrangements, and provides yet another element of a potentially unifying information-theoretic framework for complex networks.

The selection of contributions shows that there are many open research avenues in ALife that involve complex networks, such as evolvability, artificial chemistries, social networks, game theory, genetic regulatory networks, and information theory. We hope that the papers in this special issue will further motivate the cross-fertilization between the study of complex networks and ALife.

\section*{Acknowledgements}

We should like to thank all the reviewers for this special issue for their timely responses and useful comments:
Chris Adami, Lee Altenberg, Alain Barrat, Randall Beer, Hugues Bersini, Johan Bollen, Markus Brede, Mikhail Burtsev, Alan Dorin, Nic Geard, Carlos Gershenson, Mario Giacobini, Juan Luis Jiménez Laredo, Joseph Lizier, Michael Mayer, Juan Julián Merelo Guervós, Oliver Obst, Charles Ofria, Mahendrarajah Piraveenan, Mikhail Prokopenko, Tom Ray,  Hiroki Sayama, Hideaki Suzuki, Vito Trianni, Elio Tuci, Rosalind Wang, Borys Wrobel, and Larry Yaeger. We are also grateful for the support provided by the organizers of the ALife XII conference for facilitating the organization of the session on complex networks. We also appreciate the effort of all authors who submitted to the special issue and/or the ALife XII session.

\bibliographystyle{alj}

\bibliography{add,carlos,sos,complex,rbn,evolution}

\end{document}